\documentclass[conference]{IEEEtran}
\IEEEoverridecommandlockouts
\usepackage{cite}
\usepackage{amsmath,amssymb,amsfonts}
\usepackage{algorithmic}
\usepackage{graphicx}
\usepackage{textcomp}
\usepackage{xcolor}
\usepackage{array}
\usepackage{enumitem}
\usepackage{hyperref}
\setlist[itemize]{itemsep=0.5em}
\def\BibTeX{{\rm B\kern-.05em{\sc i\kern-.025em b}\kern-.08em
    T\kern-.1667em\lower.7ex\hbox{E}\kern-.125emX}}
\begin{document}

\title{Money Recognition for the Visually Impaired: A Case Study on Sri Lankan Banknotes\\}

\author{\IEEEauthorblockN{Akshaan Bandara}
\IEEEauthorblockA{\textit{Department of Software Engineering} \\
\textit{Informatics Institute of Technology}\\
Colombo, Sri Lanka \\
akshaanbandara@gmail.com}
}

\maketitle

\begin{abstract}
Currency note recognition is a critical accessibility need for blind individuals, as identifying banknotes accurately can impact their independence and security in financial transactions. Several traditional and technological initiatives have been taken to date. Nevertheless, these approaches are less user-friendly and have made it more challenging for blind people to identify banknotes. This research proposes a user-friendly stand-alone system for the identification of Sri Lankan currency notes. A custom-created dataset of images of Sri Lankan currency notes was used to fine-tune an EfficientDet model. The currency note recognition model achieved 0.9847 AP on the validation dataset and performs exceptionally well in real-world scenarios. The high accuracy and the intuitive interface have enabled blind individuals to quickly and accurately identify currency denominations, ultimately encouraging accessibility and independence.
\end{abstract}


\section{Introduction}
A disability is a condition that hinders an individual's ability to perform certain tasks or interact with others. The most common types of disabilities are Vision Impairment, Speech Impairment, and Cognitive Impairment. The World Health Organization (WHO) estimates that at least 1.3 billion of the world's population suffer from an impairment \cite{b1}. In Sri Lanka, approximately 1.6 million people are disabled \cite{b2}.

Vision impairment is the most common disability among known disabilities. According to WHO, at least 295 million individuals suffer from severe visual impairment. Moreover, at least 43 million of them are blind \cite{b3}. It is estimated that 3.9 million people in Sri Lanka suffer vision loss, of which 89,000 are blind \cite{b4}.

In Sri Lanka, banknotes are used for the majority of financial transactions. A normal person has no difficulty identifying banknotes. However, it is challenging for those with vision impairments to identify banknotes. As a result, people frequently take advantage of their disability to defraud them. This issue needs to be given more focus. It is, however, a matter that receives minimal attention.

A thorough literature review was conducted on current initiatives and research in Sri Lanka and other countries. The physical approaches include embossed dots on the banknotes. But, those dots tend to wear off. Moreover, some research has already been conducted using Deep Learning (DL) and Machine Learning (ML). The solutions proposed in those works can accurately identify the banknotes. However, the majority of the work is not user-friendly. Hence, it is necessary to solve these constraints. The goals of this study are established based on those limitations.

Thus, the research aims to design, develop, and evaluate a user-friendly interface that will allow Visually Impaired People (VIP) to accurately and efficiently identify Sri Lankan money (specifically banknotes) in diverse backgrounds.

Since Sri Lanka is a developing country, many blind people, especially in the rural areas, might not have access to smartphones. Additionally, those blind people don't often use physical currency. However, VIP in urban areas use both physical currency and smartphones for their daily activities. So, VIP who utilize cash and smartphones for their everyday activities will be the intended audience for the proposed system. As a result, the most precise solution will be a mobile application. More precisely, an application that can both identify Sri Lankan banknotes in real time and function as a stand-alone system should be developed. Thus, an efficient and accurate Object Detection (OD) model could be utilized on a dataset that has varying lighting and background images to achieve this task. Since there is not much image data for the Sri Lankan banknotes, data augmentation techniques could be used to expand the image dataset. Eventually, it should improve the accuracy of the model.

The rest of the paper is as follows: Section II reviews the previous work and outlines the research gap. A detailed description of the methodology is provided in Section III. The results are presented in Section IV, and the paper concludes with an overview of future work in Section V.
\section{Literature review}
\subsection{Related work}
The most common way VIP use banknotes is by placing those in various parts of their wallets. A relative does this task on behalf of the individual. Thus, it is convenient for the blind person to handle banknotes. Nonetheless, the visually impaired individual should always depend on others to put the money in the appropriate places.

\begin{figure}[htbp]
    \centerline{\includegraphics[width=0.45\textwidth]{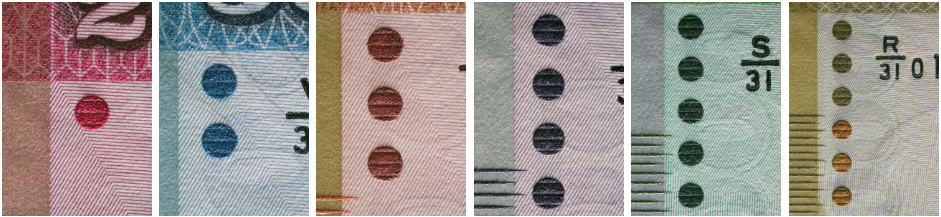}}
    \caption{Heavily printed dots on Sri Lankan banknotes \cite{b5}.}
    \label{fig:my_image_1}
\end{figure}

Furthermore, the Central Bank of Sri Lanka (CBSL) has taken initiatives to aid VIP in identifying denominations. So, as illustrated in Fig.~\ref{fig:my_image_1}, heavily printed dots appear on the left side of the currency note.

Although it is an important step towards helping the VIP, the heavily printed dots wear out with use. As a result, the VIP finds it increasingly difficult to use banknotes.

Due to recent advancements in Artificial Intelligence (AI), some excellent work has been done to address this issue. Nevertheless, more research needs to be done on Sri Lankan banknotes. Hence, the work that has been done in Sri Lanka and other countries is taken into consideration. Furthermore, only the work done in the last five years was considered for the literature survey of this research.

A study on the Sri Lankan notes and coins is presented in \cite{b6}. Convolutional Neural network (CNN) was used to classify the denominations. The note detection accuracy was 0.6667 and the coin detection accuracy was 0.6970. The low accuracy might have occurred due to CNN training from scratch. However, in their approach, an image should be captured and then the inference should be done. Thus, it can lead to a less user-friendly application.

Another research on the Sri Lankan banknotes is presented in \cite{b7}. A CNN-based Transfer Learning (TL) model was fine-tuned to classify the Sri Lankan banknotes. Popular pre-trained models such as MobileNetV2, NASNetMobile, EfficientNetB0, DenseNet121, DenseNet201, ResNet50V2, VGG16, and VGG19 were trained and tested to identify the best-performing model. As a result, DenseNet121 performed well by achieving 98\% of the test dataset. A focus mechanism is used to recognize the denominations. Hence, the model will detect the currency note when it is approximately 10 cm from the camera. The solution is less user-friendly from the perspective of a blind person, despite the model's exceptional accuracy.

There has not been much research done on this area in recent times in Sri Lanka. Thus, research done in other countries is also considered to derive the research gap.

A study was done on the Bangladeshi banknotes utlizing the Oriented FAST and Rotated BRIEF (ORB) algorithm \cite{b8}. OpenCV was used to generate the key points, which were then stored in a database. In production, when an image is received, it is converted to grayscale to eliminate the lighting conditions. The ORB keypoint detection algorithm was then used to identify the key points, and the ORB descriptor extractor was used to extract them. The brute force Hamming algorithm was then utilized to match the key points. The highest number of matching points is finally measured. The average matching time was 0.17s, and the accuracy was 100\%. This approach is, however, not a user-friendly solution.

Another research was done for the Pakistani currency notes using DL \cite{b9}. The dataset used to train the AlexNet was enhanced by incorporating data augmentation techniques. The system receives a video with the currency note in it. The trained model then identifies the banknotes in the frames after they have been separated. The approach achieved a testing accuracy of 96.85\%. Yet, this method is time-consuming.

A research was conducted to identify Yemeni currency using MobileNet \cite{b10}. A collected dataset was used to train the model. Additionally, the collection included notes in a variety of formats, including damaged and worn notes. The testing accuracy was 100\%. The model identifies the correct denomination after capturing an image of the currency note. The deployed model performs efficiently since MobileNet was utilized as the pre-trained model. However, this method is not real-time, and VIP may find it more difficult.

The literature review of the above research concluded that neither of the works is based on real-time approaches. Thus, research based on OD algorithms was reviewed. However, no research has been done on Sri Lankan currency notes incorporating OD algorithms. Conversely, OD approaches have been used in research in other nations to identify banknotes in real-time.

A study on Iraqi banknotes has leveraged the popular YOLOv3 (You Only Look Once) algorithm to train its model \cite{b11}. The dataset was created manually, and geometric transformations were applied to augment it. The images were then annotated using the LabelImg tool. The mean average precision (mAP) was 97.405\%.

A study was done on the Indian currency notes utilizing the YOLOv3 algorithm \cite{b12}. Data augmentation was done on a self-built dataset, and it was manually annotated to train the model. The standalone system has a detection and recognition accuracy of 95.71\% and 100\%, respectively. Furthermore, the system is robust enough to recognize even in cases of partial occlusions as well as wrinkled or torn currency notes.

A research done on the Ethiopian currency stands out from all the research done to date \cite{b13}. The YOLOv5m was trained on a dataset created after data augmentation and manual annotation. The model achieved an mAP of 97.9\%. Moreover, the total loss of the model was 0.044. The model performed well for the testing data by achieving a 92\% accuracy. In this work, not only variants of the YOLOv5 model were trained and tested, but also Faster\_RCNN\_Inception\_v2 and SSD\_Mobilenet\_v2 models were trained and tested. Yet, YOLOv5m outperformed the other models.

\subsection{Research gap}
The proposed system is based on OD techniques. Thus, it will allow real-time currency note recognition. Moreover, the proposed solution focuses on training and testing a model that could be easily deployed to a smartphone. To the best of knowledge, this is the first Sri Lankan currency recognition system to identify currency notes in one stage.

\vspace{0.15cm}
\section{Methodology}
The proposed solution:
\begin{itemize}
    \item Should provide accurate results despite the way the banknotes are focused.
    \item Should perform well in varying backgrounds.
    \item Should provide results in real-time.
    \item Should be easily deployable to a smartphone.
    \item Should provide results without an internet connection.
\end{itemize}

\begin{figure}[htbp]
    \centerline{\includegraphics[width=0.5\textwidth, height=5cm]{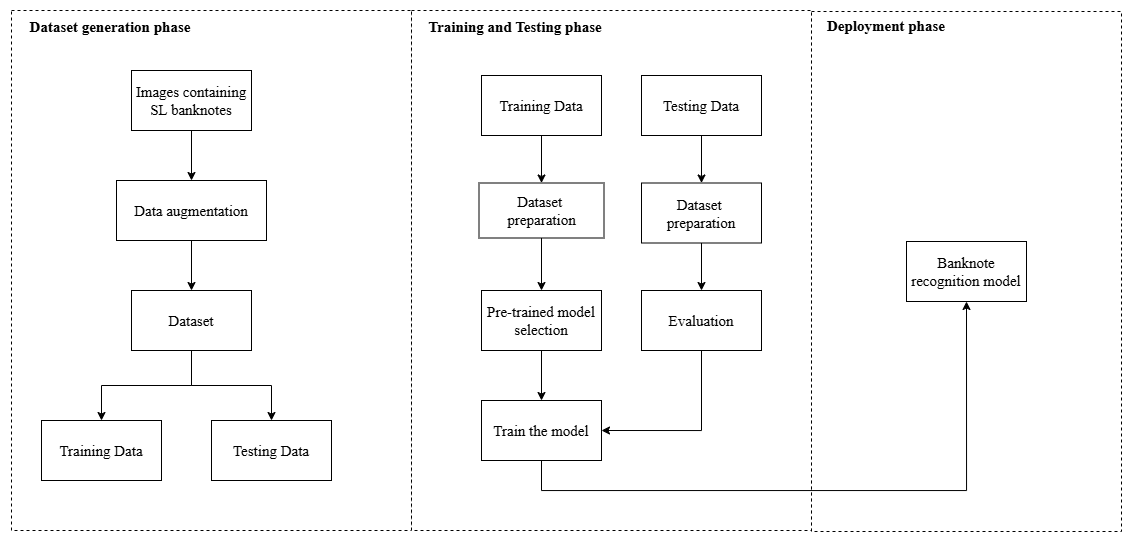}}
    \caption{Steps of the methodology.}
    \label{fig:my_image_2}
\end{figure}

Thus, this research will focus on addressing the above aspects. The steps of the methodology is presented in Fig.~\ref{fig:my_image_2}.

\vspace{-0.3cm}
\subsection{Data Gathering}
The Sri Lankan banknotes were captured using a smartphone in order to minimize data drift. If the image were captured through a digital camera, there is a high chance that the model would not perform well for a mobile phone camera feed.

There are various series of currency notes currently circulating in Sri Lanka. However, the proposed system was only trained to identify Rs 20, Rs 50, Rs 100, Rs 500, Rs 1000, and Rs 5000 of the eleventh series of currency notes issued in 2010 \cite{b14}.

\begin{figure}[htbp]
    \centerline{\includegraphics[width=0.4\textwidth, height=4cm]{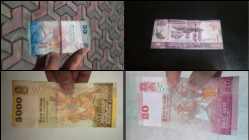}}
    \caption{Sample of captured images.}
    \label{fig:my_image_3}
\end{figure}

It is vital to equally capture images of both the front and back sides of each denomination. It is also important to capture banknotes in varying backgrounds and lighting conditions. The proposed solution should provide accurate results and perform well in varying backgrounds. A sample of captured Sri Lankan banknote images is illustrated in Fig.~\ref{fig:my_image_3}.
\begin{figure}[htbp]
    \centerline{\includegraphics[width=0.35\textwidth, height=3.5cm]{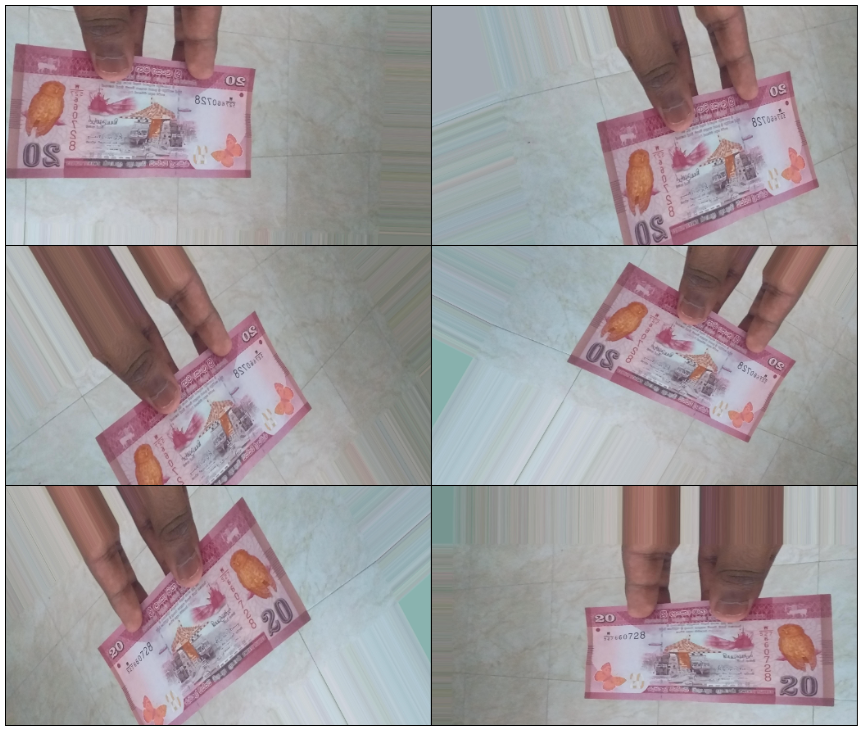}}
    \caption{Sample of generated images.}
    \label{fig:my_image_4}
\end{figure}
\subsection{Dataset Generation}
The image dataset was enhanced through data augmentation techniques. More precisely, techniques such as random rotations, random flipping, random shearing transformations, random zooming, etc. were used to enhance the banknote dataset. A sample of Sri Lankan banknote images after applying data augmentation techniques is shown in Fig.~\ref{fig:my_image_4}.
\subsection{Dataset Preparation}
The banknote recognition is an OD task. So, the currency notes in the images should be annotated. The LabelImg tool was utlized to label the generated data in the Pascal VOC format. The classes are `20 Rupees', `50 Rupees', `100 Rupees', `500 Rupees', `1000 Rupees', and `5000 Rupees'.
\begin{figure}[htbp]
    \centerline{\includegraphics[width=0.4\textwidth, height=4cm]{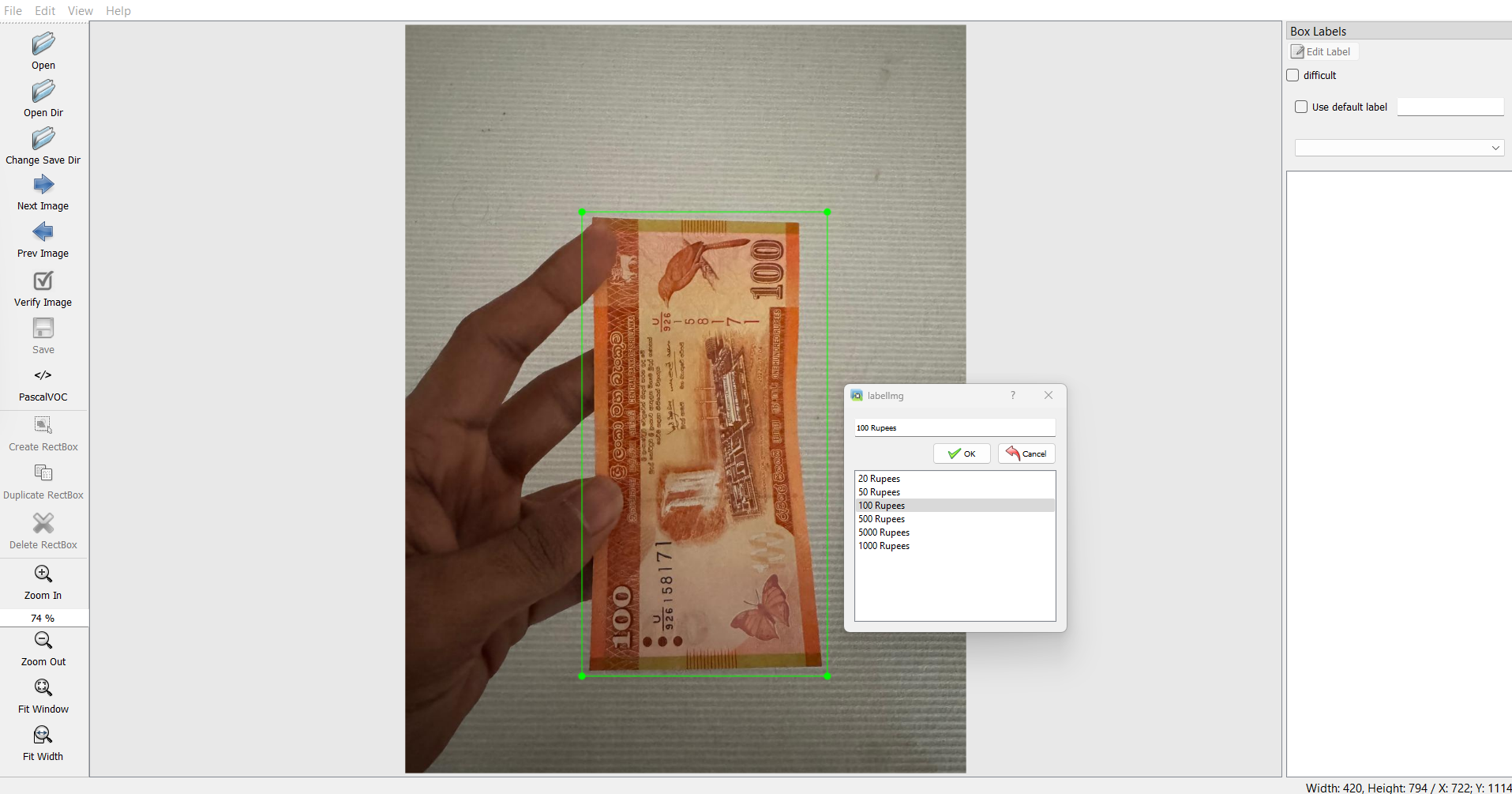}}
    \caption{Annotation of the image dataset.}
    \label{fig:my_image_5}
\end{figure}
\subsection{Pre-trained model selection}
\begin{figure}[htbp]
    \centerline{\includegraphics[width=0.5\textwidth, height=3cm]{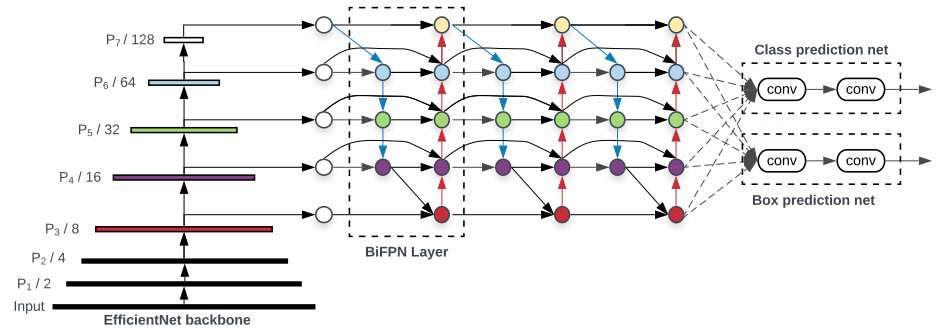}}
    \caption{Architecture of EfficientDet \cite{b21}.}
    \label{fig:my_image_6}
\end{figure}
The goal of this phase is to select a pre-trained model that is both accurate and efficient. The latest OD algorithms such as Yolov11 \cite{b15}, Yolov10 \cite{b16}, RT-DETR \cite{b17}, RT-DETRv2 \cite{b18}, Co-DETR \cite{b19}, RTMDet \cite{b20}, and EfficientDet were reviewed to identify the best-fit algorithm. As a result, EfficientDet was selected as the pre-trained model for the banknote recognition task. In EfficientDet, the accuracy and efficiency were improved by utilizing a weighted bidirectional feature pyramid network (BiFPN) and a customized compound scaling method \cite{b21}. The EfficientNet serves as the backbone network and BiFPN is the feature network in EfficientDet. Fig.~\ref{fig:my_image_6} illustrates the architecture of the EfficientDet.
\subsection{Training the model}
The banknote recognition model must be in the TensorFlow Lite (.tflite) format to be deployed to the mobile application. As a result, banknotes can be recognized without an internet connection. The accuracy and inference speed will remain unaffected even if the model size is reduced to operate on a low computationally powered space. Thus, lite versions of EfficientDet were utilized as the pre-trained model.

EfficientDet provides mobile-size lite models such as EfficientDet-lite0, EfficientDet-lite1, EfficientDet-lite2, EfficientDet-lite3, EfficientDet-lite3x, and EfficientDet-lite4. Table~\ref{tab:efficientdet_performance} presents the performance metrics for each of the pre-trained models.
\vspace{-0.2cm}
\begin{table}[htbp]
    \caption{Performance and Latency of Various EfficientDet-lite Models \cite{b22}}
    \resizebox{\columnwidth}{!}{
        \begin{tabular}{|>{\raggedright\arraybackslash}p{3cm}|c|c|c|c|}
            \hline
            \textbf{Model} & \textbf{mAP (float)} & \textbf{Quantized mAP (int8)} & \textbf{Parameters} & \textbf{Mobile Latency} \\
            \hline
            EfficientDet-lite0 & 26.41 & 26.10 & 3.2M & 36ms \\
            EfficientDet-lite1 & 31.50 & 31.12 & 4.2M & 49ms \\
            EfficientDet-lite2 & 35.06 & 34.69 & 5.3M & 69ms \\
            EfficientDet-lite3 & 38.77 & 38.42 & 8.4M & 116ms \\
            EfficientDet-lite3x & 42.64 & 41.87 & 9.3M & 208ms \\
            EfficientDet-lite4 & 43.18 & 42.83 & 15.1M & 260ms \\
            \hline
        \end{tabular}%
    }
    \label{tab:efficientdet_performance}
\end{table}

The objective was to select a pre-trained model that has a good trade-off between accuracy and mobile latency. Thus, \textbf{EfficientDet-lite2} was selected as the pre-trained model.

\subsection{Model deployment}
The banknote recognition model was deployed into the mobile application. As a result, the identification of the currency notes could be performed without an internet connection.

\subsection{Voice-based output}
The voice-based output was produced utilizing Android's TextToSpeech (tts) library. Thus, the class that is output will be read aloud for the blind individual.

\subsection{The proposed solution}
\begin{figure}[htbp]
    \centerline{\includegraphics[width=0.43\textwidth, height=3.5cm]{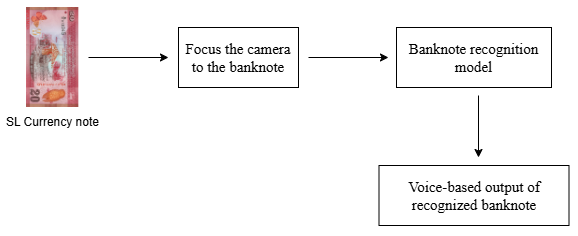}}
    \caption{The workflow of the proposed system.}
    \label{fig:my_image_7}
\end{figure}
Once the mobile application is opened, the blind person will be notified by a welcome message. The blind individual will then be able to point banknotes to the camera. The standalone banknote recognition model will identify the currency note, and the system will provide a voice-based output. The workflow of the proposed system is illustrated in the Fig.~\ref{fig:my_image_7}.

\section{Results}
The training was done on free Google Colab Graphics Processing Unit (GPU). The model was trained for 20 Epochs, and the Batch size was 4.

\subsection{Model Performance}
The performance of the OD model is evaluated using several metrics, such as \textbf{Average Precision (AP)} and \textbf{Mean Average Precision (mAP)}, which are widely utilized to evaluate the performance of the OD algorithms.

The AP is the area under the precision-recall curve. Thus, it summarizes the precision-recall curve to a single value. For AP to be high, both precision and recall should be high. Equation \eqref{eq_3} is used to calculate the value of AP.
\begin{equation}
AP = \int_0^1 P(R) \, dR \label{eq_3}
\end{equation}

\begin{itemize}
    \item \( P \): Precision, the proportion of true positives predictions among all predicted positives instances.
    \item \( R \): Recall, the proportion of true positive predictions among all actual positive instances.
    \item \( P(R) \): Precision at recall, the precision corresponding to a specific recall value.
\end{itemize}
\vspace{0.2cm}
The mAP is the average of the AP across all the classes. Equation \eqref{eq_4} is used to calculate the value of mAP.
\begin{equation}
mAP = \frac{1}{N} \sum_{i=1}^{N} \text{AP}_i \label{eq_4}
\end{equation}

\begin{itemize}
    \item \( N \): Total number of classes.
    \item \( \text{AP}_i \): Average Precision for class \( i \).
\end{itemize}
\begin{table}[htbp]
\caption{Overall performance of the model}
\begin{center}
        \begin{tabular}{|>{\raggedright\arraybackslash}p{3cm}|c|c|c|c|}
            \hline
            \textbf{Model} & \textbf{AP} & \textbf{AP$_{50}$} & \textbf{AP$_{75}$} & \textbf{mAP} \\
            \hline
            EfficientDet-lite2 & 0.9847 & 0.9983 & 0.9983 & 0.9877 \\
            \hline
        \end{tabular}
    \label{tab:moneyrec_performance_1}
\end{center}
\end{table}
\vspace{-0.5cm}
\begin{table}[htbp]
\caption{Performance of the model for each denomination}
\begin{center}
        \begin{tabular}{|>{\centering\arraybackslash}p{3cm}|>{\centering\arraybackslash}p{2.5cm}|}
            \hline
            \textbf{Denomination} & \textbf{Result} \\
            \hline
            AP (20 Rupees) & 0.9893 \\
            \hline
            AP (50 Rupees) & 0.9790 \\
            \hline
            AP (100 Rupees) & 0.9906 \\
            \hline
            AP (500 Rupees) & 0.9807 \\
            \hline
            AP (1000 Rupees) & 0.9907 \\
            \hline
            AP (5000 Rupees) & 0.9778 \\
            \hline
        \end{tabular}
    \label{tab:moneyrec_performance_2}
\end{center}
\end{table}

The currency note recognition model performed exceptionally well on the validation dataset. The results are illustrated in Table~\ref{tab:moneyrec_performance_1} and Table~\ref{tab:moneyrec_performance_2}.

\begin{figure}[htbp]
    \centerline{\includegraphics[width=0.5\textwidth, height=5cm]{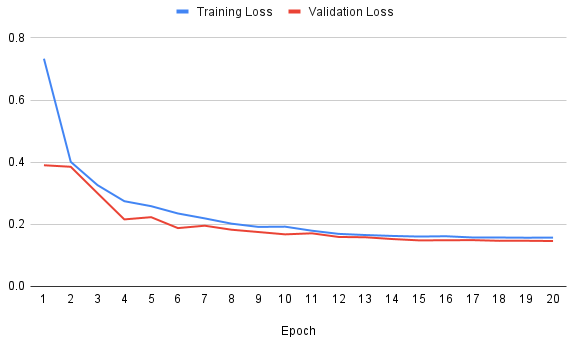}}
    \caption{Training and validation loss curves.}
    \label{fig:my_image_8}
\end{figure}

The graph in Fig.~\ref{fig:my_image_8} illustrates that the model is learning effectively and capturing important patterns in the data since there is a rapid decrease in both training and validation loss within the initial epochs. As a result, the model had learnt the majority of the features needed to perform the currency note recognition task by the 10\textsuperscript{th} epoch. The training loss and the validation loss are almost closely aligned. It implies that the model has good generalization and does not overfit. Furthermore, the consistent decrease in both losses suggests that the model is not underfitted. Eventually, in the 20\textsuperscript{th} epoch both the losses reached the optimal state.

The final training and validation loss is depicted in Table~\ref{tab:moneyrec_performance_4}. Moreover, the remaining final performance loss metrics are described in Table~\ref{tab:moneyrec_performance_5}.
\begin{table}[htbp]
    \caption{Training and Validation Loss Metrics}
    \resizebox{\columnwidth}{!}{
        \begin{tabular}{|>{\raggedright\arraybackslash}p{3cm}|c|c|c|c|}
            \hline
            \textbf{Model} & \textbf{Training Total Loss} & \textbf{Validation Total Loss} \\
            \hline
            EfficientDet-lite2 & 0.1572 & 0.1462 \\
            \hline
        \end{tabular}
    }
    \label{tab:moneyrec_performance_4}
\end{table}
\vspace{-0.5cm}
\begin{table}[htbp]
    \caption{Performance Loss Metrics}
    \resizebox{\columnwidth}{!}{
        \begin{tabular}{|>{\centering\arraybackslash}p{4cm}|>{\centering\arraybackslash}p{3.5cm}|>{\centering\arraybackslash}p{3.5cm}|}
            \hline
            \textbf{Metrics} & \textbf{Training Result} & \textbf{Validation Result} \\
            \hline
            Detection Loss & 0.0791 & 0.0681 \\
            \hline
            Classification Loss & 0.0691 & 0.0621 \\
            \hline
            Bounding Box Loss & 0.0002 & 0.0001 \\
            \hline
        \end{tabular}
    }
    \label{tab:moneyrec_performance_5}
\end{table}

\subsection{Performance in Real-World}
The developed system performs exceptionally well in the real-world, providing accurate and efficient results for the VIP in various backgrounds. Moreover, the user will receive an appropriate message if no currency notes have been identified. For a demonstration of the system’s real-world performance, see the video demo available at the \textbf{\href{https://youtu.be/BIuL7MnUzo4}{link}}.

\section{Conclusion and Future Work}
This research presents a system that can accurately and efficiently identify Sri Lankan currency notes, outperforming the existing techniques. The results indicate that the model achieved \textbf{0.9847 AP} and demonstrated robust performance in the real-world. Thus, the system will aid the visually impaired community in Sri Lanka in identifying banknotes and has the potential to make a meaningful impact on their daily lives.

This work does have limitations like any other research. It is difficult to identify banknotes at night because the Android mobile application does not allow the flashlight to be turned on while the camera is in use. Additionally, during testing, a small delay in the prediction was observed, which was caused by the video feed that is being transformed into frames for prediction. These issues are among the constraints when developing Android mobile applications.
Future research could examine the newest frameworks for mobile application development, such as Flutter, React Native, or Ionic, which might provide solutions to these limitations. Additionally, adding the capability to identify currency notes from other nations would improve the system's value and make it a tool that is relevant worldwide.

In conclusion, this study represents a significant step forward in helping the VIP identify Sri Lankan currency notes. This work establishes a foundation for future research in inclusive and accessible technology by addressing its current limitations.

\end{document}